\documentclass[11pt]{article}
\usepackage[utf8]{inputenc}
\usepackage{amsmath}
\usepackage{booktabs}

\usepackage[letterpaper]{geometry}
\usepackage{graphicx}
\usepackage{xcolor}
\usepackage[english,colorlinks=true,linkcolor={blue},citecolor={blue}]{hyperref}

\usepackage{booktabs}
\usepackage{authblk}

\usepackage[
backend=biber,
style=apa,
]{biblatex}

\DeclareCiteCommand{\cite}[\mkbibparens]
  {\usebibmacro{prenote}}
  {\usebibmacro{citeindex}%
   \printtext[bibhyperref]{\usebibmacro{cite}}}
  {\multicitedelim}
  {\usebibmacro{postnote}}

\DeclareCiteCommand*{\cite}[\mkbibparens]
  {\usebibmacro{prenote}}
  {\usebibmacro{citeindex}%
   \printtext[bibhyperref]{\usebibmacro{citeyear}}}
  {\multicitedelim}
  {\usebibmacro{postnote}}
  
\title{Representation and Interpretation in Artificial and Natural Computing}
\author[1,*]{Luis A. Pineda}
\affil[1]{Universidad Nacional Aut\'onoma de M\'exico, IIMAS, Mexico City, 04510, Mexico}
\affil[*]{lpineda@unam.mx}
\date{}

\begin{document}
\maketitle

\begin{abstract}
    \noindent
Computing is a relational phenomenon between an objective process that transforms representations mechanically and a subjective interpretation process performed by humans; in artificial computing the machine and the interpreter are different entities, but in the putative natural computing both processes are performed by the same agent. The method or process that transforms a representation is called here \emph{the mode of computing}. The mode used by digital computers is the algorithmic one, but there are others, such as quantum computers and diverse forms of non-conventional computing, and there is an open-ended set of representational formats and modes that could be used in artificial and natural computing. A mode may achieve feats beyond what the Turing Machine does, and be more powerful than such machine in a particular  sense,  but the devices would not be of the same kind and could not be compared. For a mode of computing to be more powerful than the algorithmic one, it ought to compute functions lacking an effective algorithm, and Church Thesis would not hold. Here, a thought experiment including a computational demon using a hypothetical mode for such an effect is presented. If there is natural computing, there is a mode of natural computing whose properties may be causal to the phenomenological experience. Discovering it would come with solving the hard problem of consciousness; but if it turns out that such a mode does not exist, there is no such thing as natural computing, and the mind is not a computational process.
\end{abstract}


\flushbottom

\section{Intuitive notions of computing machines}
\label{sec:introduction}

The Turing Machine (TM) was adopted as the general model of computing very soon after Turing's original paper was published \cite{turing-1936}. Together with Church's Thesis, stating that the TM computes the full set of functions that can be computed intuitively by people given enough time and material resources, or alternatively that every fully general machine or model of computing is equivalent to the TM \cite{boolos-jeffrey-1989}, defines the notion of computing or computability: that a TM does. According to this view, all digital computers are physical realization of the TM. Opposing such a strong current of opinion, one can ask whether there are other forms of computing differing genuinely from Turing's conception. In particular, the so-called ``natural computing'': If all computing engines are TMs, and \emph{the brain/mind} is a computing engine, then \emph{the brain/mind} is a TM. However, does the computer, the human invention, characterize properly a natural phenomenon that has never been observed directly? Is it not that there is here an inversion of the scientific method, that starts from observing a phenomenon, making and induction, formulating a hypothesis, that should be verified empirically? Is it not that natural computing, if there is indeed such a thing, should be characterized on its own? These questions can be addressed from the perspective of what a computer engine actually does: transforming representations. In the case of the TM, the symbols on the tape in the initial and final states represent the argument and the value of the function to be computed, respectively, and the job of the engine is to carry on with the corresponding transformation. Indeed, the difference between standard machinery and computers is that, while the former performs useful and predictable work, the only thing the latter does is transforming representations to be interpreted by humans and possible by other higher-evolved animals. However, such functionality may be achieved by various physical means, natural or artificial, opening the possibility of conceiving a large set of computing machines with their own particular properties.

\section{Representation and Interpretation}
\label{sec:rep}

A representation is a set of marks or distinctions on a physical medium --a material object-- that is interpreted as \emph{mental content}. We do not know what is the ultimate nature of mental contents, but we have introspective access to them: knowledge, beliefs, desires, intentions, feelings, emotions, pain and fear, and every thing that constitutes our psychological life, are mental contents. These are private and subjective to every cognitive individual, hence inaccessible to objective scientific investigation directly. However, they can be shared through communication, which provides a window to the mind. If there is communication, there is representation; and if there is representation, there is intentional action and interpretation. In spoken language, mental contents are ``placed on'' the wave sound, the medium, intentionally by the speaker, and are interpreted and ``placed in'' the mind of the listener. The mental contents in the minds of the speaker and hearer are similar to the extent to which communication is successful, but may differ in many respects due to contingencies of the production, transmission, and reception of the message, including the nature of the motor and sensory organs involved in linguistic communication, and the knowledge, beliefs, and expectations of the communicating individuals. Agents also interpret directly the signals and forces of the world, and in any case, the mind is constituted by interpretations. The ``transduction'' from material objects into mental contents and vice versa emerged at some point in the phylogenetic history, possibly very early, and underlies the nature of experience and consciousness, but we do not know what these are, nor why and how they appeared. These questions underlie the hard problem of consciousness, for which we do not have satisfactory answers. 

 Are there mental contents without communication? There may be animal species that perform intentional actions and make interpretations without communicating, but the window to the mind would be closed and we could not tell. Is there communication without mental content? Does the sunflower communicate with the Sun, or just senses the light that produces a chemical reaction and induces a mechanical force, and follows the sun? Does an excavator communicate with the stones that it carries or just applies a force to move them? Masses, forces, and signals constitute the material world, but are not contents. Do computers communicate? Do they make intentional actions and interpretations? Do they have mental contents, or rather, they only process signals and apply forces in the material world? It is reasonable to believe that humans and animals with a sufficiently developed neural system do make the transduction between the material and the mental, but there is no evidence to sustain that other kinds of material entity experience the world, feel that they are alive and have some form of consciousness.

Evolution provided natural representations, such as the sign systems used by animals for communicating. In the case of humans, the paradigmatic form of natural representations is spoken language. Then, at the birth of civilization appeared forms of expression with a conventional character, such as the manuscript language, which used the paper as the medium, and the ink to write on the marks, inaugurating the history of textual representations. Written language allowed communicating at different locations and times, and provided the first form of external memory for recording the facts: accounting, historical, biographical, and the literary art, and also allowed humans to make calculations that cannot be done mentally, the origin of algorithms. The second chapter was opened with the invention of movable printing types and the printing press, a machine that automates the production of representations. The third came with the invention of telecommunication typographic machines, such as the telegraph, in which texts were not only output by automatic means, but could also be typed in. The invention of the computer gave rise to the fourth chapter in which representations are not only input and output but are also transformed by a purely mechanical process, implementing algorithms. Babbage's analytical engine used mechanical gears as the medium, whose positions played the role of digits, which were interpreted as numbers. A Turing's great insight for the design of the universal computing engine was the use of typographic text as the representational format, which allowed that everything that can be expressed through text can be computed potentially. Each of the fourth chapters of the history of representations gave rise to a great cultural revolution, and we are living in the fourth era.

There is no computation without representation, and there is no representation without interpretation: if there were no agents --intentional entities capable of making intentional actions and interpretations--  ``representations'' would be inert matter, such as stones on the bed of a river wrinkled by the flow of water. Hence, there is no computation without representation and interpretation. The interpreter must know the interpretation conventions, including the notation and the standard configuration, and quite a lot of common sense knowledge to understand the configuration as a representation and not as a wrinkled stone. The interpreter must be a player in a representational language game. Artificial computing, the human invention, has an objective aspect, which is the machine that transforms the representation, and a subjective one, which is the interpretation; hence, computing is a relational objective-subjective phenomenon. Cognition is based on the hypothesis that the mind is a computing process and the brain is a computing engine. In this latter view, representations become ``internalized'', giving rise to \emph{the representational hypothesis}: the mind consists not only of interpretations, but also includes representations of interpretations. The question is what the form of such putative objects is. It is implausible that it is the typographic text, neither any alternative equally expressive external conventional format. If there are internal natural representations, they are much older in the history of life than conventional ones. Could it be that the format is the spoken language? This is indeed a natural one, but it may be just for communication, and there may be another for computation. There are also mental contents that are not linguistic, such as music, feelings, pain, fear, etc., that are experienced directly; hence they are interpretations, but their representational formats are very unlikely to be linguistic. Candidate putative formats would be those used by working memory \cite{baddeley-1981,baddeley-1992}; long-term memory, both semantic and episodic \cite{tulving-1984,tulving-2002}; the formats of the objects produced by the scene construction process  \cite{Hassabis-2007-trends}, used in navigation, imagination, vivid dreaming, etc.; or implicit memory procedural formats putatively supported by the basal ganglia, the cerebellum, and the motor cortex, among other brain structures; but we do not know their form, or whether they are indeed representations, that is, marks on a medium. Furthermore, their mere internalization does not make them intentional. Random Access Memories (RAM) of digital computers are very much like paper, and their states like ink marks! the Language of Thought (LoT) \cite{Fodor} is supposed to be intentional, but what about its interpreter? Do the expressions on the internal tape interpret themselves? If the expressions of LoT are representations, this explanation is empty. Alternatively, if the interpretation is a computational process, but the interpreter is outside the TM, there is a natural computational process, the one who understands, that is outside of the TM. Hence, there is a computing device that is not a TM. In the putative natural computing, there should be an objective machine that transforms an internal representation, an objective brain process, but also a subjective interpretation, performed by the brain of the same agent. The whole idea of representational cognition rests on finding out what is/are the representational format/s of the mind and what is the nature of the interpretation that makes representations intentional. This is again the strong problem of consciousness.

\section{Artificial Intelligence}
\label{sec:AI}

\begin{figure}[ht]
\includegraphics[width=0.4\textwidth]{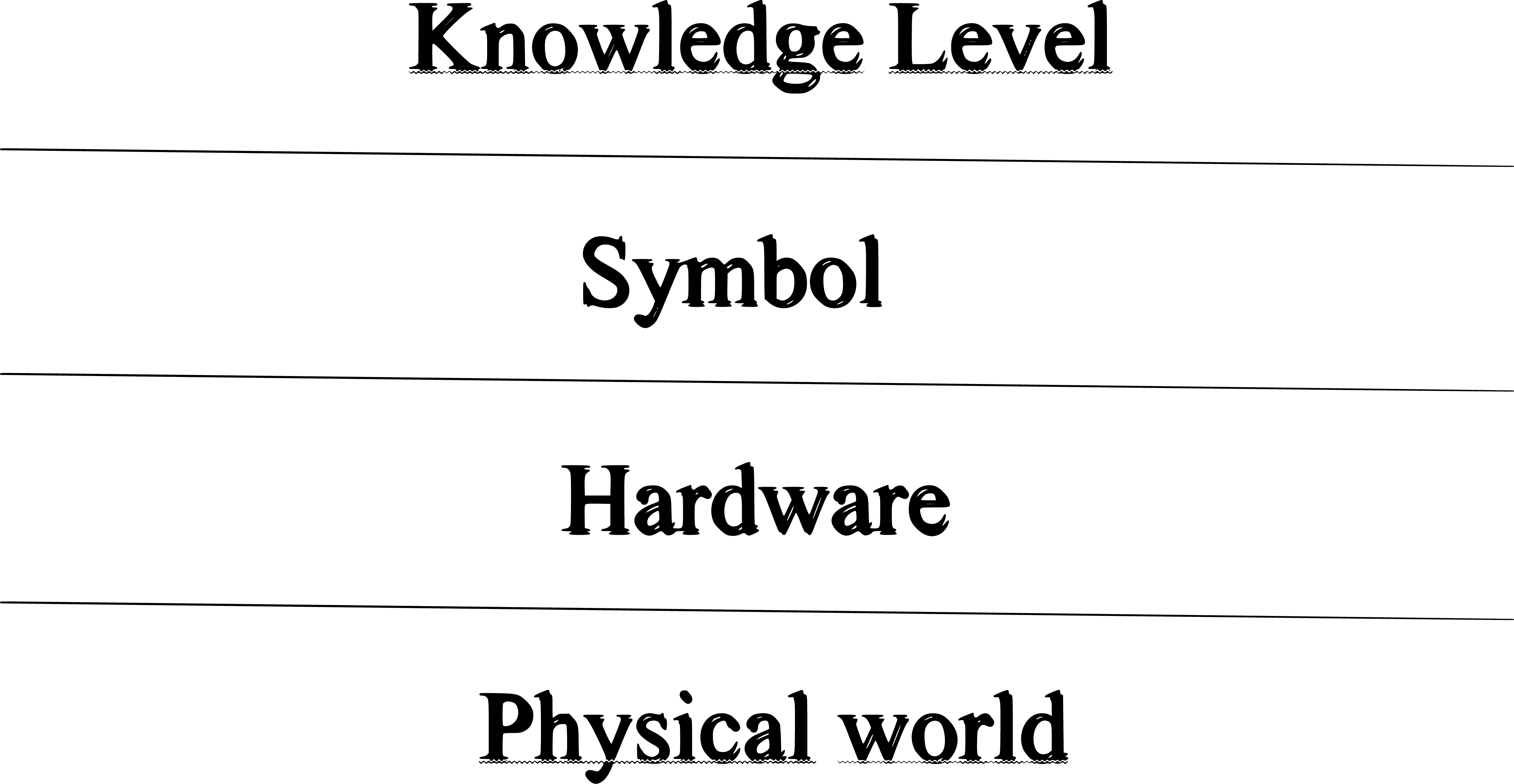}
\centering
\caption{Newell's hierarchy of system levels \cite{Pineda-2024}.}
\label{fig:knowledge-level}
\end{figure}

Turing introduced the so-called \emph{imitation game} in the 1950 paper, which was popularized as Turing's test, and stated that a machine that won it should be ascribed thought, understanding, and consciousness \cite{turing-1950}, supporting that such machines have mental contents. He also advanced the construction of intelligent machines and gave a clear illustration with his chess program Turochamp, that used extensive symbolic search. After almost three decades of research in Artificial Intelligence (AI), Simon and Newell stated the so-called Physical Symbol Hypothesis according to which a system of physical symbols --the TM-- has the necessary and sufficient conditions to generate general intelligence \cite{Newell-Simon}. Physical symbols are representations; hence, the machines that manipulate them, make interpretations, and have content. Newell went further with such a line of discussion and stated that computing machines should be analyzed at different system levels, such that each level has its own functionality, input and output, and can be studied independently of the others \cite{Newell}. He postulated a hierarchy of system levels that includes the physical world at the bottom, with several physical hardware levels on top of it, supporting the so-called \emph{symbol level}, which is directly below \emph{the knowledge level}, as illustrated in Figure \ref{fig:knowledge-level}. The knowledge and the symbol levels correspond roughly to the computational and the algorithmic levels of Marr's system levels hierarchy \cite{Marr}.

The symbol level is also the level in which software programs are stated and evaluated, and corresponds to the TM's representations, and according to Newell, the representations expressed at such a level are interpreted into the knowledge level where the medium is knowledge itself, and the only causal law is ``the principle of rationality'', giving rise to a putative computational consciousness. However, knowledge is content, and such a ``medium'' is not material; furthermore, the ``units of knowledge'' at such level are acted upon by a non-material principle; hence, Newell adopts a dualist ontology, and the claims are not scientific. However, we can reinterpret Newell's hierarchy simply by stating that the knowledge level is human knowledge, so the representations at the symbol level are interpreted by people. This move corresponds to Searle's distinction between \emph{strong} and \emph{weak} AI \cite{Searle}, the former corresponding to Turing and Newell's views and the latter to the suggested reinterpretation of Newell's hierarchy \cite{Pineda-2024}. Searle's Chiness Room mental experiment presented in the same paper shows that symbolic manipulation can be performed without understanding, so a computing entity may have a symbol level without sustaining a knowledge level.

There have been alternative proposals to Turing's intuition of computing, such as Connectionism, especially in the form of Rumelhart's parallel distributed program \cite{Rumelhart}. He claimed that cognition emerges from simple computing units assembled into large networks, such that individual units interact with their local neighbors, giving rise to distributed representations \cite{Hinton-1986} that are computed through massive parallel processes, the so-called Artificial Neural Networks (ANN). Rumelhart stated explicitly in the introduction of his book that ANNs are more powerful than TMs, challenging Church Thesis. From the present perspective, the neural network level would be placed instead of the symbol level in the hierarchy. However, although the typical diagrams illustrating the network structure and functionality --which should be placed at a system level immediately below the knowledge level too, as diagrams are representations to be interpreted-- suggest a distributed architecture and a parallel information flow, in their computer implementation the diagrams are translated into vector and matrix representations and operations, and their computation is no different from models of other conceptual domains that use the same kind of mathematical structures and algorithms, and ANN are TMs. Hence, Rumelhart's challenge is not sustained. Connectionist systems were also conceived as representational, at least in Rumelhart's formulation, and correspond to the strong AI view, and the claim that they have contents cannot be sustained neither.

Another proposal was Brooks's program of intelligence without representation \cite{Brooks,Brooks-IJCAI-1991} that gave rise to the so-called embedded cognition. This approach consisted in implementing robotic mechanisms using sensors and actuators of various kinds, either constructed physically or simulated with computers, programmed with so-called procedural representations, as opposed to declarative ones. However, if there are no representations there are neither procedural representations, and Brooks's move consists rather of placing the knowledge level directly above the hardware levels, adopting a \emph{non-representational} view of the mind. Hence, the computing process is no longer causal and essential to the robot's behavior. In practice, embedded devices are control systems that use processors contingently, such as automatic pilots. A modern car has a large number of computers embedded within its circuitry, but it is still a car. Control systems, such as thermostats and speed controllers for trains and ships, were available long before the invention of computers, do useful work and have an objective character, but do not represent anything nor transform representations, and should be considered standard machinery. The representational and non-representational views of the mind can be placed within psychological structuralism and functionalism, respectively \cite{Chemero-2013}. The two currents of thought admit interpretations --without them there would be no mind-- and the contention is whether the mind holds representations too. Brooks's robots are machines but not living entities, and the question for embedded cognition, as well as for strong AI, is how machines can have phenomenological experience. 

\section{Machines versus organisms}
\label{sec:machines}

Machines are human inventions: devices that do useful work in predictable ways. Turing stated that the determinism of digital computers is more perfect in practice than that advocated by Laplace in the 1950 paper, and computers are the paradigmatic case of determinate machines. Indeterminacy involves the entropy, which is not included in the definition nor in the functionality of the TM. Living individuals, on their part, are natural organisms produced by evolution --they are not human inventions-- and have some level of indeterminacy. Speaking of them as machines may derive from a deterministic conception of the Universe and the unity of science: the Universe is like a clock, and every thing within it is a machine too; hence there is no need to make the distinction between living entities and machines. However, rather than holding strong irreducible positions, the notion of machine may be relaxed, so the less determinate the entity, the lesser its machine-like nature. In particular, intentionality, agency and the mind, may enjoy an indetermination space where the entropy is not too low nor too high \cite{Pineda-2024}, Section 12. Low or very low indeterminacy characterizes machines and automata; conversely, if the entropy is very high, there is very little structure or none, the behavior is chaotic, and life cannot be sustained. Biological organs, such as the heart, controlled by automatic biological mechanisms, such as homeostasis, may be considered standard biological machinery, but biological structures that support intentionality should not. Indeterminacy may be a necessary condition for intentionality, but it is not sufficient because what gives rise to phenomenological experience is still unexplained. Hybrid systems integrate biological with artificial machinery, and may also integrate machinery with the biological organs supporting higher mental function and intentionality. Turing explicitly stated in the presentation of the imitation game that humans are not machines, so the question of whether machines can think has content \cite{turing-1950}. The opposition between organisms and machines is central to natural and artificial computing, and should be considered in a check list of questions to clarify the positions on machine intelligence \cite{Rouleau-2024}.

\section{The Mode of Computing}
\label{sec:mode-comp}

The are alternative formalisms to the TM, such as the Lambda calculus, the theory of Recursive Functions, and the so-called Abacus Computation, which use registers as in the Von Nuemann architecture \cite{boolos-jeffrey-1989}. Turing himself showed the equivalence between the TM and the Lambda calculus in the appendix of the 1936 paper \cite{turing-1936}, and all these machines compute the same set of functions and can be reduced to each other. Even recurrent neural networks are argued to be equivalent in this regard \cite{Sun}. All these formalisms use algorithms: a formal or mechanical procedure that transforms the representation of the argument of a function into the representation of its value, hence corresponds to the symbol level, which can be subsumed into a more general level that here we call \emph{the algorithmic level}. Turing thesis, in its mathematical sense, is based on this equivalence.

However, there are other means of transforming representations by methods or processes, natural or artificial, that do not follow an algorithm, but do so by other means, such as analogical and quantum computers, as well as other forms of non-conventional computing, such as neuromorphic systems that use memristive materials and devices \cite{Ziegler2020}. We refer to such process or methods, including the algorithmic one, as \emph{modes of computing}, and replace the symbol level in Figure \ref{fig:knowledge-level} by a generic level, which we refer to as \emph{Mode of Computing} \cite{Pineda-2024}. Particular modes require appropriate interfaces to express input and output representations and the characterization of its interpretation process. 

The mode of computing allows us to conceive other intuitive notions of computing, different from Turing's conception. All that is required is to substitute the algorithmic level by a novel mode with its corresponding representational format in the system levels hierarchy. Two main kinds of modes are the determinate and indeterminate ones. The TM is the paradigmatic case of the former, but there are modes that are not. Indeterminacy comes in two brands: that of process and that of representation. A process is indeterminate whenever there is no unique action to be performed by the computing engine at some state and symbol being scanned, so the machine's transition table is not a function but a relation. A representation is indeterminate, in turn, whenever marks on the medium that are scanned or read out have no unique interpretation. A word or a sentence may be ambiguous, but the shapes of symbols of the alphabet --the types-- are not, and textual representations are determinate. Instance of indeterminate representations are ambiguous pictures, such as the duck-rabbit image that appears in Wittgenstein's Philosophical Investigations, Gestalt ambiguous figures, or the shapes that emerge in the course of diagrammatic proofs of geometric problems, such as the Theorem of Pythagoras \cite{pineda-2007}. Their interpretation involves an active perceptual process in which the marks on the medium need to be read off in relation to a context or to a cue, either external or internal to the mind, and an interpretation act may yield more than one interpretation or none. The interpretation may even be unstable, switching from one to another in arbitrary ways. Indeterminate representations have an entropy, which is maximal whenever all possible interpretations are equally likely, and zero when only one is possible, and the representation becomes fully determinate.

A paradigmatic instance of indeterminacy of process is the so-called Probabilistic Turing Machine (PTM), in which the transition table of the standard TM is extended with probabilities such that every pair of a state and a symbol being scanned has an associated probability of performing a particular action and moving to a particular state; hence every such pair has an associated probability distribution and an entropy value. The entropy of the machine as a whole is the average entropy of all such distributions. PTMs may obey an entropy trade-off to the effect that computations are more efficient for a moderate value of the entropy, such that this is not too low nor too high. In the limits, the zero entropy PTM is a TM, and a PTM with maximal entropy is a non-deterministic TM directly. In the latter case, functions are computed through an uninformed search process and may require huge memory and time resources. Any PTM can be reduced to its equivalent TM, that computes the same function, by simply dropping the probabilities, rendering a non-deterministic TM, but computing such function may require a larger effort and may be untractable; conversely, a non-deterministic TM can be extended to a PTM by providing the probabilities explicitly, or through interacting with the environment. In any case, the TM and the PTM cannot be reduced between each other without losing information or increasing it from an external source, and constitute different modes of computing. However, the PTM does not extended the set of functions that can be computed by the TM and Church Thesis is not challenged. 

Indeterminacy of representation is exemplified by the relational-indeterminate computing, used in the Entropic Associative Memory, in which the object of computing is not the function but the relation, and \emph{relational evaluation} takes a function as its argument --which is called \emph{the cue}-- and constructs a novel function out of the cue and the indeterminate relation in a stochastic way --\emph{the recollection} \cite{pineda-eam-2021,pineda-weam-2022,pineda-imagery-eam-2023,morales2024entropicheteroassociativememory}. The main intuition of this model is that long-term memory holds indeterminate representations, in which particular remembered objects lose their identity, allowing for a very large memory capacity; however, whenever a cue is projected into the indeterminate memory mass, a novel image is constructed and retrieved as a faithful recollection, an association, or even as an imaged object. The memory has a great generalization capability but it may render false recollections, and the system as a whole obeys a memory trade-off to the effect that if the entropy is low or very low, there is a very large memory recognition precision but a very low recall; conversely, if the entropy is very high, the recall is very high but the precision drops significantly; however, there is an entropy interval, where the entropy is moderate, not too high, not too low, in which precision and recall have a good compromise and the memory is operative.

Different modes underlie different notions of computing, and some may have properties beyond what the Turing Machine does, but they would be of different kinds and comparing them would constitute a category mistake. In any case, allowing for alternatives to Turing's notion of computing invalidates the so-called strong version of Church-Turing Thesis, stating that the TM is the most powerful computing engine that can ever exist in any possible sense \cite{cop-church-turing}.

\section{The Computational Demon and Church Thesis}
\label{sec:thesis}

Church Thesis states that the TM computes the set of computable functions. In the intended sense, a function is computable if it has an effective algorithm: an algorithm that produces the value of an argument if it is defined, or a mark indicating that the function has no value for such an argument, for all the objects in the function's domain. To know this in general requires knowing whether the machine will halt or not for each of the function's arguments, and the so-called Halting Machine (HM) would have to be available. It is well known that the HM cannot be a TM \cite{boolos-jeffrey-1989}. Hence, for knowing that a function is computable, it is necessary to find an algorithm that computes it, and show that it is effective in terms of its particular structure. Church Thesis would be refuted if a single function that does not have an effective algorithm were found but computed, as the machine performing such a feat would not be a TM. 

Let us think of computing functions through a mental experiment. Suppose that there is \emph{the computational demon}. This is an omniscient being, such as Laplace's, who knows and is able to compute instantly all functions, total and partial, with finite and infinite domains and co-domains; and as Maxwell's demon, can interact with the material world, and read and write representations, so when is presented with the representation of an argument and a function, produces the representation of its value, if it is defined, or a mark indicating that such function has no value for such an argument otherwise. Let us suppose that the demon is assisted by a myriad of little computational demons, each capable to do the same feat for a particular individual function, such that every function has its little demon, so once the computational demon receives a function and its argument, handles the computation to the corresponding little demon, who performs the computation and returns the value to the master.

Let us define the set $R$ including the demons that compute computable functions. For this, we denominate the functions with finite domain and co-domain, both total and partial, as \emph{finite}, and all other functions as \emph{infinite} functions. The finite functions are enumerable through a computable function, and have a unique identifier, which also provides its extensional definition, and can be computed in a table whose columns correspond to its arguments and whose rows to their possible values \cite{Pineda-2024}, Section 11. Hence, all the little demos that compute finite functions are in $R$. The identifiers are the names of the corresponding demons, which can use any particular mode of computing to carry on with the computation, including the table, or an algorithm. That is up to the little demon. The tables help to picture that there are functions that have a clear pattern relating arguments and values systematically, and hence are structured objects, but there are others lacking such a pattern, so the relation between arguments and values is quite arbitrary.

The computational demon can also generate finite total and partial functions --included in the enumeration-- by assigning to each object in the domain an object in the co-domain, or no object if the function is partial, such that all the assignments are equally likely. In this setting, the co-domain can be thought of as a uniform probability distribution. A function generated by this procedure would have little structure or none, and would be unlikely to have an intensional definition and an algorithm. The computational demon can also generate functions with a larger degree of structure by restricting the possible values of each object in the domain to a particular subset of the co-domain, say by considering the subsets of possible values of its neighbors, reducing the entropy of the distribution. In the limit, if the subset considered for each argument has only one element, the entropy of the corresponding distribution is zero. If such an object is the same for all the arguments, the generated function is a constant function. More generally, the computational demon can assign a probability of being selected to each value in the co-domain, including the no-value case, for each object in the domain, and the entropy associated to the domain as a whole is the average entropy of the distributions associated to its arguments. It is plausible that the degree of structure of functions depends on the entropy of the distributions used in their generation, so functions generated using distributions with low or moderate entropy are very likely to have an algorithm, possibly with low computational complexity; functions generated using distributions with larger entropy are less likely to be defined and/or computed; and functions whose generation involves very high or maximal entropy are likely to be undefinable and non-computable.

In any case, a finite function lacking an algorithm can nevertheless be computed through its table or its extensional definition, and Church Thesis is trivially true for this set. This is the case for digital computers that have a finite register size. However, the enumeration function and the amount of memory required to hold the table grow exponentially, and in practice it is necessary to find algorithms that can be computed with current digital computers, or use an alternative mode, such as quantum computing. 

We now turn to the case of infinite functions. As in the finite case, there are infinite functions with a great deal of structure, which are very likely to have an intensional definition and an algorithm. A function may be anonymous but described by an algorithm directly, or the algorithm may be implicitly defined, such as those computing functions learned by current machine deep learning algorithms or produced by evolutionary computation, although strictly speaking, such functions are finite and only approximate infinite ones. An infinite function is computable by a TM if it is definable by any means and has an effective algorithm, and their corresponding demons are also included in $R$. 

Conversely, there are infinite functions with very little or no structure at all, which can be imaged but are not definable and are non-computable in an absolute sense. The computational demon can achieve the feat of choosing values equally likely among the objects included in an infinite enumerable co-domain, and assign one such value to each object of an infinite enumerable domain; hence, can generate an immeasurable number of non-computable functions, and handle them to their corresponding little demons, who are not in $R$. Infinite functions generated by such a procedure cannot be named, defined, or described by people, so we cannot ask the computational demon to compute them, and their corresponding little demos are locked up in a room, a computational limbo, that will remain closed forever.

For their part, infinite but definable functions that do not have effective algorithms, hence cannot be computed by a TM, can be computed by their corresponding little demons --using a hypothetical mode of computing-- who are included in $R$. Suppose that $\Omega$ is one such function that is computed by its little demon using a mode of computing, artificial or natural, which cannot be algorithmic. Then $R$ is larger than the set of demons that compute functions that have an effective algorithm. Hence, Church Thesis is refuted in the present thought experiment. Suppose further that there is an actual mode of computing, natural or artificial, which we call \emph{the device}, that implements the demon of $\Omega$. The discovery of such a device would refute Church Thesis in its mathematical sense.

\section{Natural Computing}
\label{sec:natural-comp}

If there is natural computing, there should be a natural mode of computing directly above the brain structures involved in cognition and immediately below the knowledge level. The definition of such mode would involve finding the format/s of mental representation/s and the physical mode/s of computing that transforms them. In contrast to artificial computing, in which the entity modifying the representation objectively and the one performing the interpretation subjectively are different, in natural computing both processes are performed by the same agent, so computing and interpreting are two aspects of the same phenomenon, and experiencing the world and being conscious are the manifestations of natural computing. The brain may use different modes for directly transforming and interpreting representations, and the phenomenological aspect of intentionality may be due to their particular properties. There could even be functions lacking an effective algorithm but having \emph{a device} used in mental processes, and the brain would be a more powerful computing organ than the Turing Machine. The question of whether there is natural computing is open to empirical investigation, and finding it out would come with solving the hard problem of consciousness. If it is ever found, the representational hypothesis of mind would be upheld. However, if it turns out that such an object does not exist, the non-representational hypothesis holds, and the mind is not a computational process, but something else, not known as yet.

\section{Funding statement}

The author acknowledges the partial support of grant PAPIIT-UNAM IN103524, México. 

\section{Conflicts of Interest statement}

The author states that there are no conflicts of interest related to this paper.

\printbibliography

\end{document}